Ibrahim Alshubaily
Xiangliang Zhang
Data Mining
15 Oct 2019

# TextCNN with Attention for Text Classification

## 1. Introduction

The vast majority of the world's textual content is unstructured, making automated classification an important task for many applications. The goal of text classification is to automatically classify the text documents into one or more predefined categories. Recently proposed simple architectures for text classification such as "Convolutional Neural Networks for Sentence Classification." by Kim, Yoon [3] showed promising results. In this paper, we propose incorporating an attention mechanism into the network to boost its performance, we also propose WordRank for vocabulary selection in order to reduce the network embedding parameters and speed up training with minimum accuracy loss. By adopting the proposed ideas TextCNN accuracy on 20News increased from 94.79 to 96.88, moreover, the number of parameters for the embedding layer can be reduced substantially with little accuracy loss by using WordRank. By using WordRank for vocabulary selection we can reduce the number of parameters by more than 5x from 7.9M to 1.5M, and the accuracy will only decrease by ~1.2% .

## 2. Related Work

Text classification algorithms such as Naive Bayes, Support Vector Machines have been widely used in text classification application, due to their abilities to learn a robust model from a relatively small dataset.

Deep learning models are very effective for text classification without the need for feature engineering. The two main deep learning architectures used in text classification are Convolutional Neural Networks (CNN) and Recurrent Neural Networks (RNN).

Deep learning models require a large dataset for effective learning. Transfer learning is often applied to offset this requirement. Word embedding is one of the most popular representations of document vocabulary. It is capable of capturing context of a word in a document, semantic and syntactic similarity, relation with other words, etc.

By using a pre trained word embedding model, we can enable our network to solve the task with high accuracy from a relatively small dataset.

Using a CNN to classify text was first presented in the paper Convolutional Neural Networks for Sentence Classification by Yoon Kim [3]. The main idea is to see our documents as images. Let us say we have a sentence and we have maxlen = 70 and embedding size = 300. We can represent this sentence with a matrix of numbers with the shape 70x300 to.

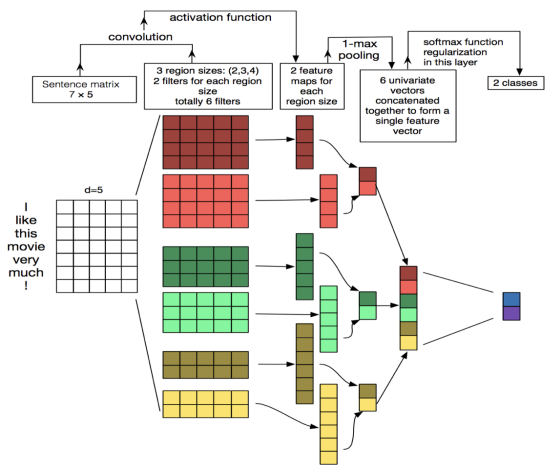

Figure 1.

Dzmitry Bahdanau et al first presented attention in their paper Neural Machine Translation by Jointly Learning to Align and Translate [1], In essence, attention is a mechanism for computing the score/importance of the inputs to determine which are more useful for solving the task at hand. Traditionally this idea is used for sequential models as demonstrated by Vaswani, Ashish, et al. "Attention Is All You Need." [6], we adopt this technique for TextCNN [3] to give the network more degrees of freedom to assign a score to the different convolutional layers/branches of the model.

## 3. Methods

In this section, I will describe the proposed additions to TextCNN and the vocabulary selection method WordRank.

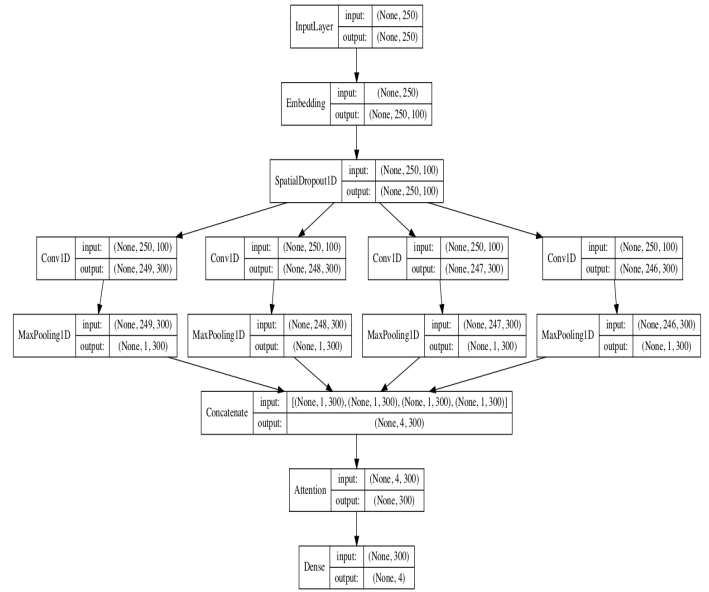

Figure 2.

### 3.1 Spatial Dropout

Dropout regularization is a computationally cheap way to regularize a deep neural network. Dropout works by probabilistically removing, or "dropping out," inputs to a layer, which may be input variables in the data sample or activations from a previous layer. It has the effect of making nodes in the network generally more robust to the inputs.

We improve the model of [textcnn] by adding an additional spatial dropout layer [5] on the embedding of the input sentence. The role of dropout is to improve generalization per-formance by preventing activations from becoming strongly correlated, which in turn leads to overtraining.

### 3.2 Pooling

Pooling layers provide an approach to down sampling feature maps by summarizing the presence of features in patches of the feature map. Two common pooling methods are average pooling and max pooling that summarize the average presence of a feature and the most activated presence of a feature respectively. As illustrated in figure 2, we apply Max pooling across the spatial dimension for the different convolutional branches to insure the output shape is the same across all outputs.

### 3.3 Attention

Applying attention [6] to the outputs of the convolutional layers served two purposes, it gave us insight about what filter sizes where more effective for the text classification task, which enabled to fine tune that parameter. For example, we tried large filter sizes (ie. 12) and ended up eliminating that branch because it had a low attention score, and therefore it was somewhat a waste of parameters. More significantly, attention gives the network the freedom of assigning a score for each of the branches which boosted the model performance. The attention mechanism resulted in a higher accuracy compared to simply flattening or averaging the output of the convolutional layers.

### 3.4 WordRank

PageRank is an algorithm for ranking web pages developed by Google. It's used to give each page a relative score of importance by evaluating the quality and quantity of its links. [2]

Power iteration is a simple iterative algorithm and calculate the PageRank score of a page without knowing the value of other pages that link to it. Each time we run the calculation, we are getting a closer estimate of the final value, we repeat the calculations until convergence which is not guaranteed in all cases.

In order to prevent some pages from having too much influence, the PageRank formula also uses a dampening factor. In the formula, the total value of pages is damped down by multiplying it by 0.85.

Applying PageRank to rank words requires a very simple transformation, we can simply treat words as web pages, and word to word similarity as the link/vote value. The similarity is calculated based on the pretrained Glove 100 dimensional word embedding. By experimenting with cosine similarity and word mover distance similarity, we found the word mover distance similarity more effective for preserving the model accuracy after reducing the vocabulary size.

Its important to point out that the number words WordRank can rank is limited, because the similarity matrix size is $|V|^2$. Therefore, it's necessary to do some preprocessing first such as removing stop words and applying limmitization in addition to removing very infrequent words.

### 4. Results

The results of the experiment shows that the proposed improvements to the TextCNN model architecture increased accuracy on 20News from 94.79 to 96.88. Moreover, the number of parameters for the embedding layer can be reduced substantially with little accuracy loss by using WordRank. For example, we can choose the vocabulary size to be 10k, which will reduce the number of parameters by more than 5x from 7.9M to 1.5M, and the accuracy will only decrease by ~1.2% .

| Model | 20News | Number of Params | Average Seconds per epoch |
|---|---|---|---|
| Bow + LR | 92.81 | - | - |
| Bigram + LR | 93.12 | - | - |
| Average Embedding | 89.39 | 7.4 M | 18 seconds |
| TextCNN | 94.79 | 7.8 M | 62 seconds |
| RCNN | 96.49 | 7.6 M | 92 seconds |
| TextCNN w Attention | **96.88** | 7.9 M | 65 seconds |

By comparing WordRank to the standard frequency based vocabulary selection, we can observe that WordRank is a more effective vocabulary selection method in terms of minimizing the validation accuracy loss.

Baseline:

| Model | 20News | |V| = 75K Number of Params |
|---|---|---|
| TextCNN w Attention | 96.88 | 7.9 M |

Comparison of vocabulary selection methods:

| Vocabulary Selection Method | |V| = 30K<br><br># Params = 3.5M | |V| = 20K<br><br># Params = 2.5M | |V| = 10K<br><br># Params = 1.5M |
|---|---|---|---|
| Frequency | 95.36 | 94.61 | 94.02 |
| WordRank | 96.20 | 96.03 | 95.62 |

## 5. Conclusion

This paper presented the effectiveness of incorporating an attention mechanism to TextCNN[3] , which resulted in a significant increase in the model accuracy and proven to be helpful for hyperparameter selection (filter size). Moreover, the paper also demonstrated how applying page rank for vocabulary selection (WordRank) is a more effective method compared the commonly used frequency based selection criteria.

To further advance the work done here, one can merge the model proposed by Lai, Siwei, et al. "Recurrent Convolutional Neural Networks for Text Classification." [4] to see if combining the two architectures can produce a more effective feature extraction method for text classification.

PageRank is an effective but somewhat outdated ranking algorithm, it is also limited in terms of the number of words it can rank due to the huge size of the similarity matrix. More advanced learning to rank models such as RankNet and LambdaRank are likely to produce more impressive results.

**References**

1. Bahdanau, Dzmitry, et al. "Neural Machine Translation by Jointly Learning to Align and Translate." *ArXiv.org*, NIPS 2014, May 2014, https://arxiv.org/abs/1409.0473 .

2. "Google's PageRank Algorithm: Explained and Tested." *Google's PageRank Algorithm: Explained and Tested*, https://www.link-assistant.com/news/google-page-rank-2019.html .

3. Kim, Yoon. "Convolutional Neural Networks for Sentence Classification." *ACL Anthology*, ACL Anthology, Oct. 2014, https://www.aclweb.org/anthology/D14-1181/.

4. Lai, Siwei, et al. "Recurrent Convolutional Neural Networks for Text Classification." *ACM Digital Library*, AAAI Press, Jan. 2015, https://dl.acm.org/citation.cfm?id=2886636.

5. Tompson, Jonathan, et al. "Efficient Object Localization Using Convolutional Networks." *ArXiv.org*, CVPR 2015, 9 June 2015, https://arxiv.org/abs/1411.4280.

6. Vaswani, Ashish, et al. "Attention Is All You Need." *ArXiv.org*, NIPS 2017, 6 Dec. 2017, https://arxiv.org/abs/1706.03762.